%% file: main.tex
\documentclass[10pt,twocolumn,letterpaper]{article}

\usepackage[pagenumbers]{cvpr}  

\usepackage{cuted}
\usepackage{graphicx}
\usepackage{capt-of}
\usepackage{xcolor}
\usepackage{xspace}
\usepackage{amsmath,amssymb,amsbsy,amsfonts,dsfont,pifont,bm,bbm,mathrsfs,mathtools,nicefrac}
\usepackage{float,wrapfig}
\usepackage{booktabs,multirow,adjustbox,diagbox,threeparttable,tabularray}
\usepackage{algorithm,algpseudocode,listings}
\usepackage{enumitem}
\usepackage[accsupp]{axessibility}
\definecolor{citeblue}{RGB}{48,111,186}
\usepackage[pagebackref=false,breaklinks=true,colorlinks=true,citecolor=citeblue,bookmarks=false]{hyperref}

\usepackage{cleveref}  
\crefname{section}{Sec.}{Secs.}
\Crefname{section}{Section}{Sections}
\crefname{table}{Tab.}{Tabs.}
\Crefname{table}{Table}{Tables}
\crefname{figure}{Fig.}{Figs.}
\Crefname{figure}{Figure}{Figures}
\crefname{equation}{Eq.}{Eqs.}
\Crefname{equation}{Equation}{Equations}
\newcommand{\R}{\mathbb{R}}      
\renewcommand{\c}{{\rm\bf c}}    
\renewcommand{\t}{{\rm\bf t}}    
\newcommand{\z}{{\rm\bf z}}      
\newcommand{\Z}{\mathcal{Z}}     
\newcommand{\w}{{\rm\bf w}}      
\newcommand{\W}{\mathcal{W}}     
\newcommand{\f}{{\rm\bf f}}      
\newcommand{\x}{{\rm\bf x}}      

\newcommand{\method}{\textcolor{black}{\mbox{\texttt{Aurora}}}\xspace}

\def\cvprPaperID{****}
\def\confName{CVPR}
\def\confYear{2023}

\newcommand\nonumfootnote[1]{%
\begingroup%
    \renewcommand\thefootnote{}\footnote{\hspace{-3.7pt}#1}%
    \addtocounter{footnote}{-1}%
\endgroup%
}

\begin{document}

\title{Exploring Sparse MoE in GANs for Text-conditioned Image Synthesis}

\author{
    Jiapeng Zhu\textsuperscript{1,2$\dagger$} \quad
    Ceyuan Yang\textsuperscript{3$\dagger$} \quad
    Kecheng Zheng\textsuperscript{1} \quad
    Yinghao Xu\textsuperscript{1,4} \quad
    Zifan Shi\textsuperscript{1,2} \quad
    Yujun Shen\textsuperscript{1} \\[5pt]
    \textsuperscript{1}Ant Group \qquad
    \textsuperscript{2}HKUST \qquad
    \textsuperscript{3}Shanghai AI Laboratory \qquad
    \textsuperscript{4}CUHK
}

\twocolumn[{
\renewcommand\twocolumn[1][]{#1}
\maketitle
\begin{center}
    \vspace{-15pt}
    \includegraphics[width=1.0\linewidth]{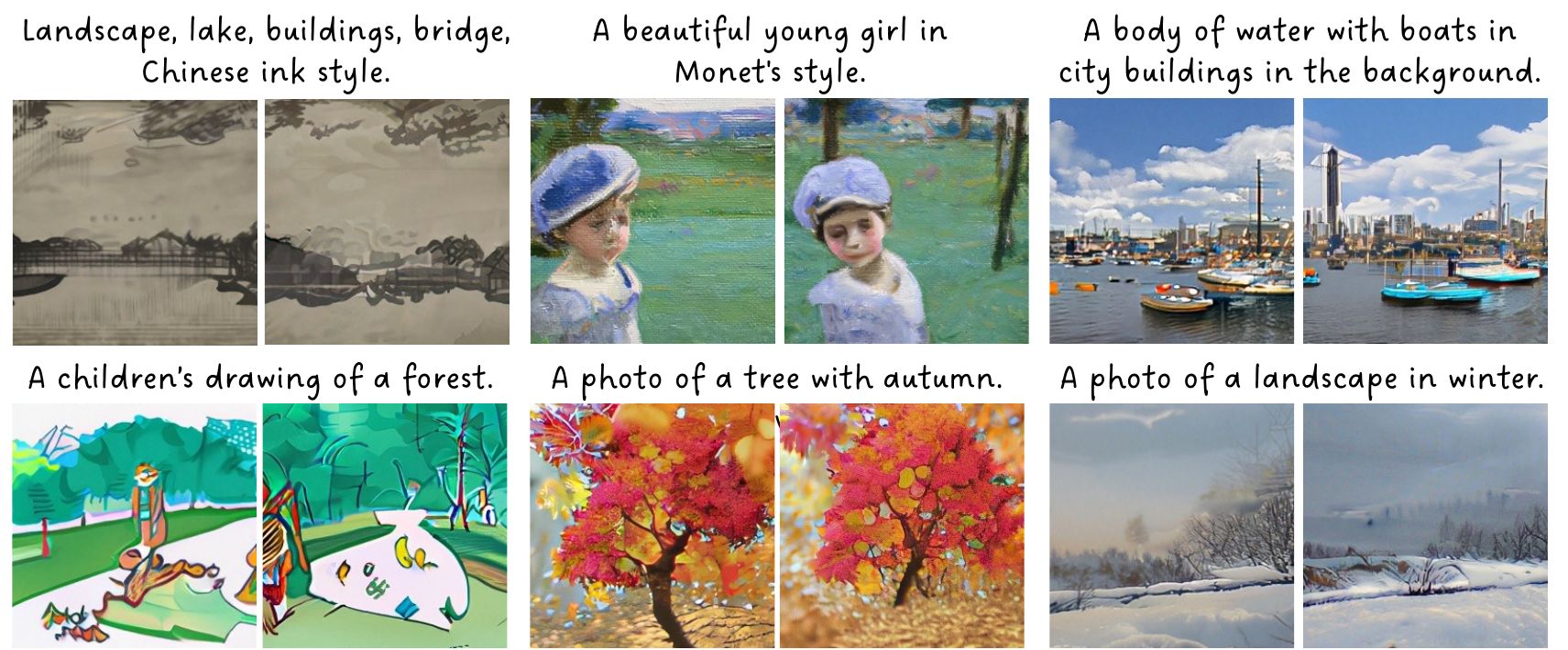}
    \vspace{-16pt}
    \captionsetup{type=figure}
    \caption{%
        \textbf{Example results} synthesized by our proposed \method, a large-scale GAN-based text-to-image generator.
        Note that the current version of our generator only produces images at $64\times64$ resolution, followed by a super-resolution model~\cite{ldm} for $4\times$ upsampling.
        We have carefully prepared a long-term plan for the improvement of \method from the aspects of both performance and functionality, which can be positively expected to directly output a high-quality synthesis without any post-processing.
    }
    \label{fig:teaser}
    \vspace{5pt}
\end{center}
}]

\input{sections/0.abs.tex}
\input{sections/1.intro.tex}
\input{sections/2.related.tex}
\input{sections/3.method.tex}
\input{sections/4.exp.tex}
\input{sections/5.conclusion.tex}
\input{sections/6.ref.tex}

\end{document}

%% file: sections/0.abs.tex
\begin{abstract}

Due to the difficulty in scaling up, generative adversarial networks (GANs) seem to be falling from grace on the task of text-conditioned image synthesis.
Sparsely-activated mixture-of-experts (MoE) has recently been demonstrated as a valid solution to training large-scale models with limited computational resources.
Inspired by such a philosophy, we present \method, a GAN-based text-to-image generator that employs a collection of experts to learn feature processing, together with a sparse router to help select the most suitable expert for each feature point.
To faithfully decode the sampling stochasticity and the text condition to the final synthesis, our router adaptively makes its decision by taking into account the text-integrated global latent code. 
At 64$\times$64 image resolution, our model trained on LAION2B-en and COYO-700M achieves 6.2 zero-shot FID on MS COCO.
We release the code and checkpoints \href{https://github.com/zhujiapeng/Aurora}{here} to facilitate the community for further development.
\nonumfootnote{$\dagger$ denotes equal contribution.}

\end{abstract}
\vspace{-20pt}

%% file: sections/1.intro.tex
\section{Introduction}\label{sec:intro}

As one of the most popular tasks in the field of content generation, text-conditioned image synthesis~\cite{dalle, dalle2, imagen, parti, ldm, gigagan} allows users to customize a picture with natural language instructions.
Among diverse types of generative models, such as variational autoencoders (VAEs)~\cite{vae}, autoregressive models (ARs)~\cite{vqvae, pixelcnn}, and generative adversarial networks (GANs)~\cite{gan}, diffusion models~\cite{sohl2015deep, ddpm, guidediffusion} have blossomed into a dominant solution with a thriving community built upon stable diffusion~\cite{ldm}, leaving few efforts on improving other frameworks.

In this work, we put our focus on GANs, which are usually criticized for training instability~\cite{wgan, sngan} and limited mode coverage~\cite{salimans2016improved}.
However, we argue that the task of text-to-image (T2I) generation is more like a translation task (\textit{i.e.}, shifting an informative data distribution to another) instead of a pure generation task (\textit{i.e.}, modeling the observed distribution with random noise as the input), and accordingly that the introduction of the text condition can already help alleviate the above two issues to some extent, which is also empirically supported by some recent studies~\cite{stylegan-t, gigagan}.
Even so, the difficulty in scaling up the model size still manifests itself as a major drawback of GANs.
Concretely, unlike diffusion models that produce an image through an iterative process~\cite{ddpm, guidediffusion, ldm}, the generator of a GAN typically adopts a feedforward network to render an image from a sampled latent code.
Considering the limited computational resources (\textit{e.g.}, GPU memory) in practice, directly learning a sufficiently large feedforward model remains challenging.

The advent of sparsely-activated mixture-of-experts (MoE)~\cite{shazeer2017outrageously,fedus2022switch} brings with it a promising way to enlarge the model scale with affordable computing cost.
The rationale behind is to employ multiple expert modules yet only activate one or a few of them for each feedforward execution.
That way, the growth of the number of modules could expand the model capacity but barely increase the number of computations.

We decently introduce the philosophy of sparse MoE into the design of GAN generators and propose \method for fast T2I generation.
Specifically, given an intermediate feature map at a particular level of degree, we engage a set of experts, each of which is originally designed to process a single-point feature, as well as a highly sparse router to help select the most suitable expert for each feature point.
It is noteworthy that, different from most existing spare MoE methods~\cite{jacobs1991adaptive, shazeer2017outrageously, fedus2022switch} that determine the routing path simply based on the information from the input feature map, our router also takes into account the text-integrated global latent code such that it can adaptively make the best decision via fully considering the sampling stochasticity and the text condition.
We progressively increase the image resolution in the training phase, and our generator learned on LAION2B-en~\cite{laion} and COYO-700M~\cite{coyo} achieves 6.2 zero-shot FID~\cite{fid} on MS COCO~\cite{coco} at the resolution of 64$\times$64.

Besides the fast inference speed, GANs also enjoy a well-studied latent space~\cite{interfacegan, higan, gansteerability} and the high flexibility to incorporate other priors (\textit{e.g.}, involving the volumetric rendering pipeline for 3D-aware image synthesis)~\cite{hologan, graf, stylenerf, stylesdf, geod, volumegan, voxgraf, epigraf, eg3d, 3dsurvey, pigan, pof3d, get3d}.
However, the lack of an open-sourced large-scale GAN model hinders such a powerful tool from being adequately developed and analyzed.
We thus release the code and checkpoints to facilitate the research community and plan to continuously improve the released model from the perspectives of both performance and functionality.
We believe that our work could encourage more studies in the field of visual content generation.

%% file: sections/2.related.tex
\section{Related Work}\label{sec:related}

\noindent\textbf{Generative adversarial networks (GANs).}
Formulated as a two-player game between a generator and a discriminator, GANs~\cite{gan} are designed to model the observed data distribution through adversarial training.
Thanks to the flexible formulation (\textit{i.e.}, there are no particular assumptions on the data format or on the generation process), GANs have shown great potential in a variety of visual generative tasks, such as image generation~\cite{pggan, stylegan, stylegan2, stylegan3, biggan}, image editing~\cite{interfacegan, zhu2016generative, higan, shen2021closed, ganspace, gansteerability, suzuki2018spatially, xu2021generative, wu2020stylespace, bau2019gandissection,zhu2021lowrankgan, zhu2022resefa}, video generation~\cite{saito2017tgan, tulyakov2018mocogan}, and 3D-aware image synthesis~\cite{hologan, graf, pigan, stylenerf, volumegan, eg3d, 3dsurvey}.
Some studies also explore conditional GANs~\cite{conditionalgan}, which incorporate signals like class labels~\cite{biggan, sagan, stylegan-xl}, texts~\cite{lafite, galip, stylegan-t, gigagan}, and reference images~\cite{isola2017image, yu2018generative, ledig2017photo} into the generator for interactive control.
Despite the aforementioned success, GANs usually struggle in learning from large-scale, diverse datasets (\textit{e.g.}, LAION~\cite{laion}) due to the difficulty in scaling up~\cite{gigagan}.

\noindent\textbf{Text-conditioned image synthesis.}
With the aid of visual-language models like CLIP~\cite{clip}, text-to-image (T2I) generation~\cite{dalle, dalle2, imagen, muse, parti, ldm, ediff,  gigagan} manages to decode an open-vocabulary description to a vivid image, and hence allows users to visualize their creativity and imagination.
Towards such a challenging task, existing efforts usually opt to trade computing efficiency for visual quality.
For example, autoregressive models~\cite{dalle, makeascene, cogview, parti} and diffusion models~\cite{glide, dalle2, ldm, imagen} have demonstrated impressive results through an iterative generation process, which can be time-consuming and costly to make a synthesis.
Differently, GANs employ a feedforward network as the generator and thus naturally support fast generation without having to use an acceleration~\cite{ddim, dpmsolver} or distillation~\cite{SalimansH22, consistency} algorithm.
Some recent attempts~\cite{stylegan-t, gigagan} have confirmed the promise of GANs in text-to-image synthesis yet left the issue of enlarging the model capacity unsolved.
In addition, there currently lacks a publicly available GAN-based T2I model in the community, making it difficult to fully study the properties of GANs, let alone integrate the advantages of GANs into other types of generative models.

\noindent\textbf{Mixture-of-experts (MoE).}
Sharing a similar philosophy with ensemble~\cite{OpitzM99,Rokach10}, MoE~\cite{jacobs1991adaptive} learns a set of expert modules to process the input signal, together with a gating module (also known as a router) to help combine the outputs of different experts.
Under such a design, each expert is expected to focus on the learning of a sub-region of the input space, while the router is responsible for the weight assignment across experts~\cite{eigen2013learning}.
Sparsely-activated MoE~\cite{shazeer2017outrageously, fedus2022switch, du2022glam} distinguishes itself from other variants of MoE because of the discrete routing mechanism.
By selecting only one or a few experts from a large candidate pool for each incoming example, sparse MoE offers a feasible solution to scaling up the model size with barely increased computational cost~\cite{fedus2022switch, riquelme2021scaling, shen2023scaling}.
In this work, we introduce the idea of sparse MoE into the design of a GAN generator and adequately adapt the routing mechanism for the task of text-to-image generation by taking into account the text-integrated stochasticity.

%% file: sections/3.method.tex
\section{Method}\label{sec:method}

Recall that we aim to develop an efficient text-to-image (T2I) generator based on the framework of GANs~\cite{gan}, which consists of a generator and a discriminator competing with each other.
On the one hand, the generator takes in a global latent code, $\z \in \R^{512}$, and a text description (\textit{e.g.}, a sentence), $\c$, and outputs an image, $\x=G(\z, \c)$.
On the other hand, the discriminator takes in an image together with an associated text and differentiates synthesized samples from real samples using a realness score, $D(\x, \c)$.
As discussed above, the major contribution of this work is to enlarge the generator capacity; therefore, we directly borrow the discriminator architecture from the prior art~\cite{gigagan}.
In the following, we will describe our generator in detail from the aspects of text information injection (\cref{subsec:text-conditioned-sampling}), network design of a unit block (\cref{subsec:unit-generative-block}), supervision signals (\cref{subsec:learning-objectives}), and training strategy (\cref{subsec:training-strategy}).

\subsection{Text-conditioned Sampling}\label{subsec:text-conditioned-sampling}

As pointed out by StyleGAN~\cite{stylegan}, applying a learnable MLP-based mapping network to the latent code, $\z$, could result in a more disentangled latent space, $\W$, than the native latent space, $\Z$, which has been shown to facilitate semantic manipulation~\cite{interfacegan, higan, ganspace, gansteerability, xu2021generative, wu2020stylespace}.
Under the T2I setting, we choose to also feed the text condition into the mapping network, following prior arts~\cite{stylegan-t, gigagan}, to make the sampling process better aware of the text information.
In particular, we adopt CLIP~\cite{clip} as the text encoder, which converts the input text, $\c$, to a token sequence, $\t_{seq}$, as well as a global token, $\t_g$, aggregating the information of the entire sentence.
Then, we concatenate the global token onto the sampled latent code for further processing as
\begin{align}
    &\{\t_{seq}, \t_g\} = \texttt{CLIP}(\c),          \label{eq:clip} \\
    &\w = \texttt{MLP}(\texttt{concat}(\z, \t_{g})),  \label{eq:mapping}
\end{align}
where $\w$ stands for the disentangled latent code.
Here, we need to point out that, considering the task gap between vision-language alignment and T2I generation, we freeze the text encoder of CLIP and stack a few more learnable layers on top of the encoder to allow flexible tuning~\cite{gigagan}.

\subsection{Unit Generative Block}\label{subsec:unit-generative-block}

\begin{figure}[t]
    \centering
    \includegraphics[width=1.0\linewidth]{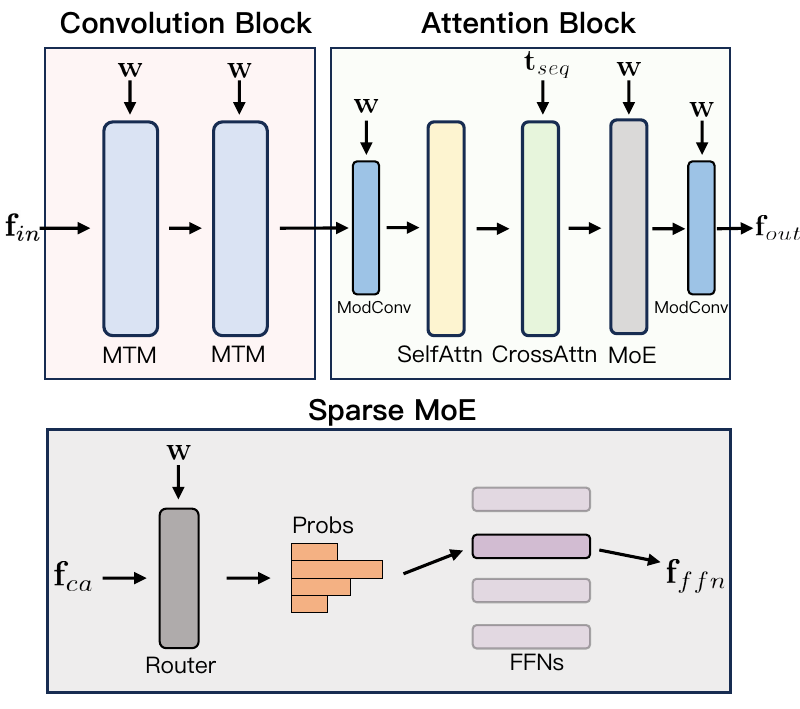}
    \vspace{-22pt}
    \caption{%
        \textbf{Illustration of a unit generative block} in our proposed \method.
        Details can be found in \cref{subsec:unit-generative-block}.
    }
    \label{fig:framework}
    \vspace{-5pt}
\end{figure}

Like most modern deep models~\cite{vaswani2017attention, stylegan}, our generator follows the principle of block-wise design.
As shown in \cref{fig:framework}, a generative unit stacks a convolution block and an attention block to process the input feature map, $\f_{in}$, where the attention block adopts the sparse routing mechanism for scaling up.

\noindent\textbf{Convolution block.}
Latent modulation~\cite{stylegan}, which is developed from adaptive instance normalization (AdaIN)~\cite{adain}, has demonstrated its effectiveness in the generation field and hence serves as a basic operation in GAN-based models~\cite{stylegan-xl, stylegan-t, gigagan}.
Such a technique is further improved through fused into the convolution operation with demodulation (\textit{i.e.}, modulated convolution, \texttt{ModConv})~\cite{stylegan2}, and through equipped with learnable feature deformation (\textit{i.e.}, modulated transformation module, \texttt{MTM})~\cite{mtm}.
We employ two \texttt{MTM}s in the convolution block and construct the block using skip connection.
Thus, we obtain the output of the convolution block as
\begin{align}
    \f_{conv} = \texttt{MTM}\Big(\texttt{MTM}(\f_{in}, \w), \w\Big) + \f_{in}. \label{eq:conv}
\end{align}
Notably, the upsampling operation is omitted for brevity, and the learnable offsets in \texttt{MTM}~\cite{mtm} are only introduced when the feature resolution is equal to or less than $16\times16$.
Besides, all convolution operations adopt sample-adaptive kernel selection~\cite{gigagan}.

\begin{figure*}[t]
    \centering
    \includegraphics[width=1.0\linewidth]{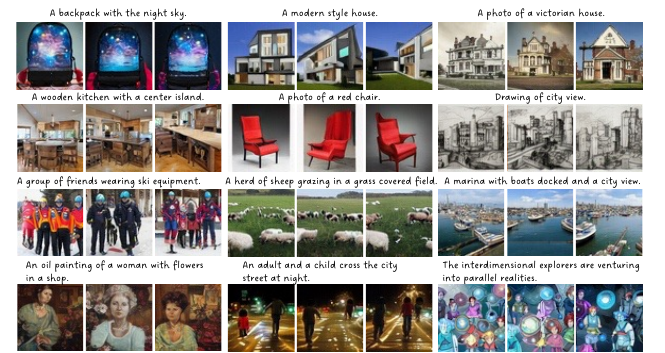}
    \vspace{-16pt}
    \caption{%
        \textbf{Diverse results} generated by \method, where we randomly sample three global latent codes, $\z$ for each text condition, $\c$.
    }
    \label{fig:synthesize-res}
    \vspace{-5pt}
\end{figure*}

\noindent\textbf{Attention block.}
We can tell from \cref{eq:conv} that, in the convolution block, the text condition, $\c$, only interacts with the generative features through the global token, $\t_g$, with $\w$ in \cref{eq:mapping} as an intermediary.
To also make sufficient use of the token sequence, $\t_{seq}$, we introduce an attention block for text-image interaction, which is widely used in multi-modality learning~\cite{LiSGJXH21, TanB19}.
Specifically, the attention block consists of a self-attention layer~\cite{vaswani2017attention}, a cross-attention layer~\cite{vaswani2017attention}, and a feedforward network~\cite{vaswani2017attention}.
We thereby have
\begin{align}
    \f_{proj} &= \texttt{ModConv}(\f_{conv}, \w),                 \label{eq:proj-in}\\
    \f_{sa}  &= \texttt{SelfAttn}(\f_{proj}) + \f_{proj},         \label{eq:self-attention} \\
    \f_{ca}  &= \texttt{CrossAttn}(\f_{sa}, \t_{seq}) + \f_{sa},  \label{eq:cross-attention} \\
    \f_{ffn} &= \texttt{FFN}(\f_{ca}) + \f_{ca},                  \label{eq:ffn} \\
    \f_{out} &= \texttt{ModConv}(\f_{ffn}, \w),                   \label{eq:proj-out}
\end{align}
where the modulated convolutions in \cref{eq:proj-in,eq:proj-out} serve as the feature projector to adjust the number of feature channels.
The \texttt{FFN} in \cref{eq:ffn} processes each feature point individually using a two-layer MLP.
Here, we use $\ell2$-distance attention~\cite{gigagan} to stabilize training.

\noindent\textbf{Multiple experts and sparse routing.}
Recall that the main motivation of this work is to enlarge the capacity of the generator in GANs.
One straightforward solution is to make the network deeper or wider. However, it may cause an out-of-memory issue or unstable training.
Inspired by the practice in natural language processing (NLP)~\cite{fedus2022switch}, we incorporate the philosophy of sparse MoE~\cite{shazeer2017outrageously} into the design of our unit generative block to enlarge the generator capacity.
Concretely, we treat the \texttt{FFN} in \cref{eq:ffn} as an expert, which specializes in processing a single-point feature, and engage $N$ experts in total, $\{\texttt{FFN}_j\}_{j=1}^N$, forming a candidate pool.
We then employ a sparse router to help select the most suitable for each feature point.
Unlike existing sparse MoE methods~\cite{shazeer2017outrageously, fedus2022switch, du2022glam}, our router adaptively makes its decision by taking into account not only the input feature but also the text-integrated global latent code, $\w$, which better matches the target of T2I generation (\textit{i.e.}, decoding the text condition and the sampling stochasticity to an image).
Under such a routing strategy, we can reformulate \cref{eq:ffn} as
\begin{align}
    \f_{ffn}^{(k)} = \texttt{FFN}_j(\f_{ca}^{(k)}) + \f_{ca}^{(k)} \ \text{s.t.}\ j = \texttt{Router}(\f_{ca}^{(k)}, \w).  \label{eq:router}
\end{align}
Here, the superscript $(k)$ denotes the $k$-th feature point.

\begin{figure*}[t]
    \centering
    \includegraphics[width=1.0\linewidth]{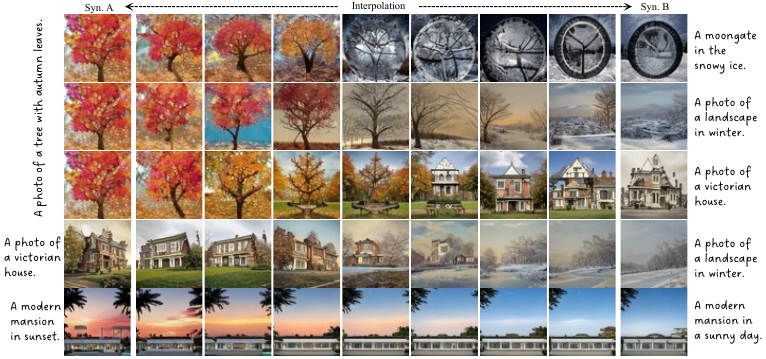}
    \vspace{-16pt}
    \caption{%
        \textbf{Synthesized results through text prompt interpolation}, where we fix the global latent code, $\z$, and interpolate the text tokens, $\{\t_{seq}, \t_g\}$, extracted from two different text conditions, $\c$.
    }
    \label{fig:prompt-interp}
    \vspace{0pt}
\end{figure*}

\subsection{Learning Objectives}\label{subsec:learning-objectives}

The primary focus of this work is to scale up the model size of a GAN generator instead of exploring a better supervision signal for GAN training.
Thus, we directly inherit some well-studied loss functions from prior arts.
Still, to make the presentation self-contained, we list below all the objectives used in the training phase.

\noindent\textbf{Adversarial loss.}
Adversarial training~\cite{gan} is the core concept shared by all GANs and owns many variants in the literature~\cite{wgan, lsgan, began}.
We follow StyleGAN~\cite{stylegan} and adopt the logistic non-saturating loss~\cite{gan} with $R_1$ penalty~\cite{mescheder2018training} as the regularizer.

\noindent\textbf{Matching-aware loss.}
Unlike the discriminator in unconditional GANs that rejects a sample by merely considering its plausibility, the discriminator in T2I GANs is also expected to take into account the alignment between the condition and the synthesized result.
We follow prior arts~\cite{reed2016generative, stackgan, gigagan} to employ a matching-aware discriminator and introduce the matching-aware loss that urges the discriminator to reject mismatched text-image pair.

\noindent\textbf{Multi-level CLIP loss.}
CLIP loss~\cite{clip} follows the idea of contrastive learning~\cite{chen2020simple}, and forces the feature of a generated image close to the feature of its associated text yet away from the feature of irrelevant text.
Such a supervision signal could complement the matching-aware loss in helping the generator better decode the text condition.
We introduce multi-level CLIP loss to encourage text-image similarity at all resolutions.

\noindent\textbf{MoE loss.}
Recall that we introduce a collection of experts to enlarge the model capacity and expect them to process the input feature in various ways.
For each network execution, only one expert will be activated for a feature point.
To balance the workload across experts (\textit{i.e.}, to avoid that some experts are never activated), we introduce the MoE loss following Switch Transformer~\cite{fedus2022switch}.

\subsection{Training Strategy}\label{subsec:training-strategy}

Considering the difficulty of open-vocabulary T2I generation, it might be challenging to directly learn the model at a relatively high image resolution.
We introduce the progressive training scheme proposed in PGGAN~\cite{pggan} and come up with an effective indicator, termed \textit{reference FID}, to help determine the timing of moving from one training stage to the next stage.
It is noteworthy that, even though our approach adopts the mechanisms of both adversarial learning and sparse MoE,  we do \textit{not} observe any instability in the entire training process.

\noindent\textbf{Progressive training.} 
Progressive training has been widely studied in the GAN literature~\cite{pggan, stylegan, pigan} thanks to its efficacy in improving training stability and efficiency.
For the task of image generation, existing studies propose to gradually increase the image resolution such that the generative blocks at lower resolution can get appropriately warmed up.
We follow PGGAN~\cite{pggan} to construct our generator from $4\times4$ resolution.
When involving a new stage, the output of the newly introduced blocks will be accumulated into the output of the previous stage to obtain the final synthesis.

\noindent\textbf{Reference FID.}
A question that naturally arises with progressive training is to find out the perfect timing to shift to a new resolution.
To enable end-to-end training with minor human participation, we propose reference FID as an automatic indicator to guide the training process.
Fréchet inception distance (FID)~\cite{heusel2017gans} is a widely used metric for quantifying the distribution similarity between a collection of real images and a collection of synthesized ones.
Instead, we independently sample two sets of real images and calculate the FID between them.
Such a score partly serves as a reference to the best performance we can expect from the generator.
We prepare the reference FID scores for each image resolution in advance and trigger the training of the next stage once the generator surpasses the reference FID at the current stage.

%% file: sections/4.exp.tex
\section{Experiments}\label{sec:exp}

In this part, we evaluate the capability of our proposed \method both qualitatively (\cref{subsec:qualitative}) and quantitatively (\cref{subsec:quantitative}), and further make detailed analyses to help readers with a better understanding of text-to-image (T2I) GANs (\cref{subsec:analyses}).
Finally, we discuss the significance of this project as well as how our open-sourced code and checkpoints could benefit the research community (\cref{subsec:discussion}).

\begin{figure*}[t]
    \centering
    \includegraphics[width=1.0\linewidth]{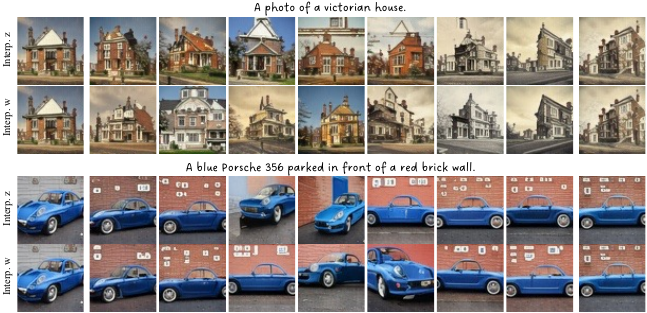}
    \vspace{-16pt}
    \caption{%
        \textbf{Synthesized results through latent code interpolation}, where we fix the text condition, $\c$, and interpolate the latent codes within the native latent space, $\Z$, and the disentangled latent space, $\W$, respectively.
    }
    \label{fig:latent-interp}
    \vspace{0pt}
\end{figure*}

\subsection{Experimental Setups}\label{subsec:setup}

\noindent\textbf{Datasets.}
We used text-image pairs from LAION2B-en~\cite{laion} and COYO-700M~\cite{coyo} to train our model, taking into account factors such as clip score~\cite{hessel2021clipscore}, image size, and aesthetic score~\cite{aesthetic2022}.
We do \textit{not} use any self-collected data to simply pursue the improvement of synthesis quality.
We believe that, with our open-sourced codebase, it would be possible to learn a more capable model with cleaner and higher-quality data.

\noindent\textbf{Training details.}
We develop our algorithm using Hammer~\cite{hammer2022} and consume 256 NVIDIA A100 GPUs for one week to learn a generator at $64\times64$ resolution.
We use AdamW~\cite{adamw} as the optimizer, with hyper-parameters $\beta_1=0$, $\beta_2=0.99$, and weight decay set to $1e^{-5}$.
The generator and the discriminator are jointly optimized with a learning rate of 0.0025.
CLIP ViT-L/14~\cite{clip} is employed as the text encoder.

\subsection{Qualitative Results}\label{subsec:qualitative}

\noindent\textbf{Text-to-image synthesis}.
We showcase some example results in \cref{fig:synthesize-res} under diverse text conditions.
We can tell that, besides performing well on image synthesis within a narrow distribution (\textit{e.g.}, human faces, horses, or bedrooms), GANs are also capable of open-vocabulary T2I generation, allowing users to play with their creativity and imagination.
It is noteworthy that our generator also manages to handle the generation of a crowded scene (see the first example in the third row of \cref{fig:synthesize-res}).

\noindent\textbf{Text interpolation.}
Our generator produces an image with two independently sampled inputs, namely, a global latent code, $\z$, and a text prompt, $\c$.
Here, we would like to investigate the properties of these two different input spaces through interpolation.
As the natural language description does not support interpolation, we choose to interpolate text prompts at the token level (\cref{eq:clip}).
We present some results in \cref{fig:prompt-interp}, where we can observe a smooth semantic transition.
For instance, when we make an interpolation between ``a tree with autumn leaves'' and ``a landscape in winter'' (see the second row of \cref{fig:prompt-interp}), the tree gradually disappears while the season harmoniously varies from autumn to winter at the same time.
Such a result suggests that our model could faithfully preserve semantic continuity when learning to decode text conditions.

\noindent\textbf{Latent interpolation.}
Latent space interpolation is widely studied in the literature~\cite{pggan, biggan, stylegan} to verify that the generator delivers the generative ability instead of memorizing the dataset.
We also perform interpolation between global latent codes, whose results are shown in \cref{fig:latent-interp}.
Here, we conduct the experiments within both the native latent space, $\Z$, and the disentangled latent space, $\W$.
Interestingly, the interpolation results are more like randomly sampled, especially taking the prompt interpolation results in \cref{fig:prompt-interp} as a comparison.
Such a phenomenon highly contradicts the commonly believed fact that GANs own a semantically continuous latent space~\cite{interfacegan, gansteerability, wu2020stylespace}.
This helps point out a future research direction, which is to disentangle the effects of text condition and sampling stochasticity.

\subsection{Quantitative Results}\label{subsec:quantitative}

Besides the visual samples exhibited in \cref{subsec:qualitative}, we also quantitatively compare our \method with existing T2I generators to give readers an overall picture of its synthesis performance.
Following prior arts~\cite{stylegan, biggan, stylegan2, stylegan-xl, stylegan-t, gigagan}, we adopt FID~\cite{fid} as the metric, which evaluates a generator at the distribution level.
In \cref{tab:comparison}, we report the evaluation results on two different data domains, including the training data domain, LAION~\cite{laion}, and the commonly used zero-shot test data domain, MS COCO~\cite{coco}.
Here, ``FID$_{10K}$'' is obtained by comparing the inception features~\cite{inceptionv3} of $10K$ synthesized images and $10K$ real images randomly sampled from the training set, while ``zero-shot $\text{FID}_{30K}$'' is obtained by comparing the inception features of $30K$ synthesized images and the full validation set of MS COCO.
We can tell that \method delivers the best results on both domains at the image resolution of $64\times64$.

\subsection{Analyses}\label{subsec:analyses}

\begin{figure}[t]
    \centering
    \includegraphics[width=1.0\linewidth]{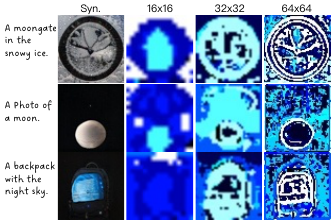}
    \vspace{-16pt}
    \caption{%
        \textbf{Analysis of our learned sparse router}, where we use different colors to visualize the index of expert selected by the router for each pixel.
        We can tell that the router tends to cluster the pixels by various visual concepts.
    }
    \label{fig:sparse-router}
    \vspace{-5pt}
\end{figure}

\noindent\textbf{Sparse router.}
Recall that our motivation is to enlarge the capacity of the generator in GANs following the philosophy of sparse MoE~\cite{jacobs1991adaptive, shazeer2017outrageously, fedus2022switch}.
Different from existing approaches, our sparse router makes its decision based on the information of not only the input feature map but also the conditional text and the sampling stochasticity (\cref{eq:router}).
Here, we visualize some routing paths to verify the effectiveness of our routing mechanism.
We can tell from \cref{fig:sparse-router} that, within a synthesized image, pixels with similar visual concepts tend to be processed by the same expert (\textit{i.e.}, having the same color in \cref{fig:sparse-router}).
Such an observation holds at all resolutions, verifying the rationality of our design.

\noindent\textbf{Latent spaces.}
As observed in \cref{subsec:qualitative}, the interpolation results within the latent space, either $\Z$ or $\W$, present unexpected behavior, contradicting our intuition that the latent space in GANs is typically semantically smooth.
Several reasons may contribute to such a phenomenon.
Firstly, the absence of the perceptual path length regularity (PPL)~\cite{stylegan2} might be a factor in this counter-intuition, which is specifically designed to improve smoothness in the latent space.
Secondly, the similarity computation in cross-attention (\cref{eq:cross-attention}) may cause the discontinuity between adjacent points on a feature map, such that the text token sequence, $\t_{seq}$, may dominate the generation process over the text-integrated global latent code, $\w$.
A better way of injecting text information may help alleviate this issue.
There may also be some additional factors at play, which we plan to investigate in the future.

\begin{table}[t]
\caption{%
    \textbf{Quantitative comparison} between our \method and existing text-to-image generators.
    FID~\cite{fid} is employed as the evaluation metric, where a smaller number indicates better performance.
    ``FID$_{10K}$'' reports the results on the training set, while ``Zero-Shot FID$_{30K}$'' reports the results on MS COCO~\cite{coco} without fine-tuning.
    All evaluations are performed at $64\times64$ resolution, and ``$*$'' means that we downsample the generated samples using Lanczos~\cite{stylegan-t} for a fair comparison.
}
\label{tab:comparison}
\vspace{-7pt}
\centering\small
\SetTblrInner{rowsep=1pt}       
\SetTblrInner{colsep=5.6pt}     
\begin{tblr}{
    cells={halign=c,valign=m},  
    column{1}={halign=l},       
    hline{1,2,6,7}={1-4}{},     
    hline{1,7}={1.0pt},         
    vline{2}={1-8}{},           
}
    Method                           & FID$_{10K}$  & Zero-Shot FID$_{30K}$ & Params.  \\
    eDiff-I$^*$\cite{ediff}          &  -           & 7.60                  &  9.10B   \\
    SD$^*$~\cite{ldm}                &  -           & 8.40                  &  0.94B   \\
    StyleGAN-T~\cite{stylegan-t}     &  -           & 7.30                  &  1.02B   \\
    GigaGAN~\cite{gigagan}           &  9.18        & -                     &  0.65B   \\
    \method (ours)                   &  8.28        & 6.45                  &  1.16B   
\end{tblr}
\vspace{-5pt}
\end{table}

\subsection{Discussion}\label{subsec:discussion}

The lack of an open-sourced large-scale GAN model makes it hard for the community to fully study the past favorite generative weapon, leaving some stereotypes, including both negative and positive ones, on GANs after scaling up for T2I generation.
For example, instead of the commonly believed training instability, we find that GANs can be \textit{stably optimized} after introducing text as the condition.
In addition, GANs are usually appreciated for the semantically continuous latent space, but we observe that the \textit{interpolation between prompts} better meets the expectation than the results of interpolating latent codes (see \cref{subsec:analyses}).
We believe that these findings would be of great interest to the community and hope that our release could inspire more findings and further encourage the development of a powerful and fully-featured generative model.

Besides, we would like to point out that our model is decently trained on publicly available datasets (see \cref{subsec:setup}) without fine-tuning on any self-collected data for performance improvement.
While our release is for research purposes only, we can positively expect the performance gain when deploying our approach on cleaner and higher-quality data.

%% file: sections/5.conclusion.tex
\section{Conclusion}\label{sec:conclusion}

In this work, we investigate how to expand the capacity of GANs to match the demand of open-vocabulary text-to-image generation.
Drawing lessons from sparse MoE, our proposed \method achieves on-par synthesis performance with existing diffusion-based counterparts yet enjoys a sufficiently faster inference speed.
We make the code and checkpoints public to facilitate the studies in the field of visual content creation.

%% file: sections/6.ref.tex
{\small
\bibliographystyle{ieee_fullname}
\bibliography{ref}
}